\PassOptionsToPackage{unicode}{hyperref}
\PassOptionsToPackage{hyphens}{url}
\documentclass[11pt]{article}
\usepackage{amsmath,amssymb}
\usepackage{lmodern}
\usepackage{iftex}
\ifPDFTeX
  \usepackage[T1]{fontenc}
  \usepackage[utf8]{inputenc}
  \usepackage{textcomp} 
\else 
  \usepackage{unicode-math}
  \defaultfontfeatures{Scale=MatchLowercase}
  \defaultfontfeatures[\rmfamily]{Ligatures=TeX,Scale=1}
\fi
\IfFileExists{upquote.sty}{\usepackage{upquote}}{}
\IfFileExists{microtype.sty}{
  \usepackage[]{microtype}
  \UseMicrotypeSet[protrusion]{basicmath} 
}{}
\makeatletter
\@ifundefined{KOMAClassName}{
  \IfFileExists{parskip.sty}{%
    \usepackage{parskip}
  }{
    \setlength{\parindent}{0pt}
    \setlength{\parskip}{6pt plus 2pt minus 1pt}}
}{
  \KOMAoptions{parskip=half}}
\makeatother
\usepackage{xcolor}
\ifLuaTeX
  \usepackage{selnolig}  
\fi
\IfFileExists{bookmark.sty}{\usepackage{bookmark}}{\usepackage{hyperref}}
\IfFileExists{xurl.sty}{\usepackage{xurl}}{} 
\hypersetup{
  hidelinks,
  pdfcreator={LaTeX via pandoc}}

\author{Axel Schaffland}
\date{}
\title{Rephotography in the Digital Era\\
	{\large Mass Rephotography and \rp, the Web Portal for Rephotography}}

\usepackage{axelPaper}
\usepackage[top=3cm,left=3cm,right=4cm,bottom=3cm]{geometry}
\usepackage{float}
\graphicspath{{Bilder/}}

\renewcommand{\rp}{re.photos\xspace}

\begin{document}
\maketitle

\hypertarget{abstract}{%
\section{Abstract}\label{abstract}}

Since the beginning of rephotography in the middle of the 19th century, techniques in registration, conservation, presentation, and sharing of rephotographs have come a long way. Here, we will present existing digital approaches to rephotography and discuss future approaches and requirements for digital mass rephotography. We present \rp\footnote{\url{https://www.re.photos}}, an existing web portal for rephotography, featuring methods for collaborative rephotography, interactive image registration, as well as retrieval, organization, and sharing of rephotographs. For mass rephotography additional requirements must be met. Batches of template images and rephotographs must be handled simultaneously, image registration must be automated, and intuitive smartphone apps for rephotography must be available. Long–term storage with persistent identifiers, automatic or mass georeferencing, as well as gamification and social media integration are further requirements we will discuss in this paper. 

\section{Where We Stand Now}
\label{where-we-stand-now}

\begin{figure}[htb]
	\centering
	\includegraphics[width=\textwidth]{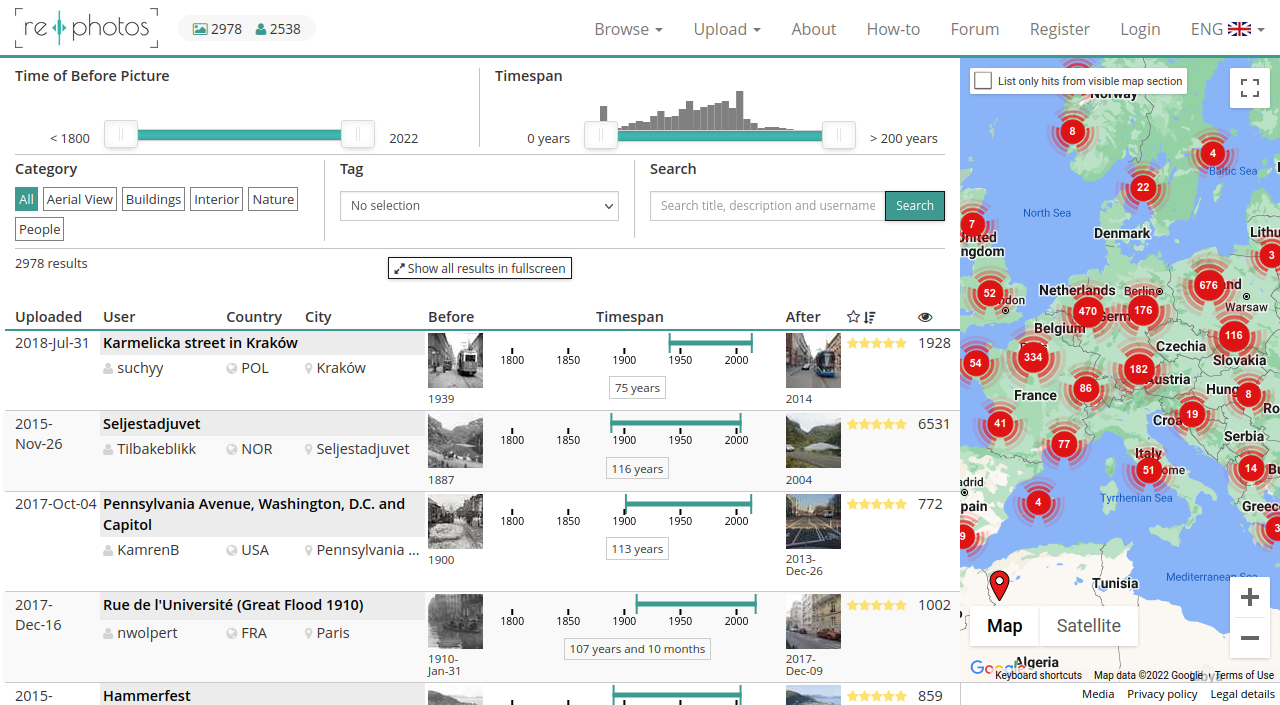}
	\caption{Browsing view of  \rp. Compilations are displayed as a list and 
	on a map. They can be filtered and sorted by different meta information,
	\eg, date, timespan, tags, titles, and descriptions or constrained to 
	a map section selectable on the right.}
	\label{fig:port2}	
\end{figure}

\begin{figure}[htb]
	\center
	\includegraphics[width=\textwidth]{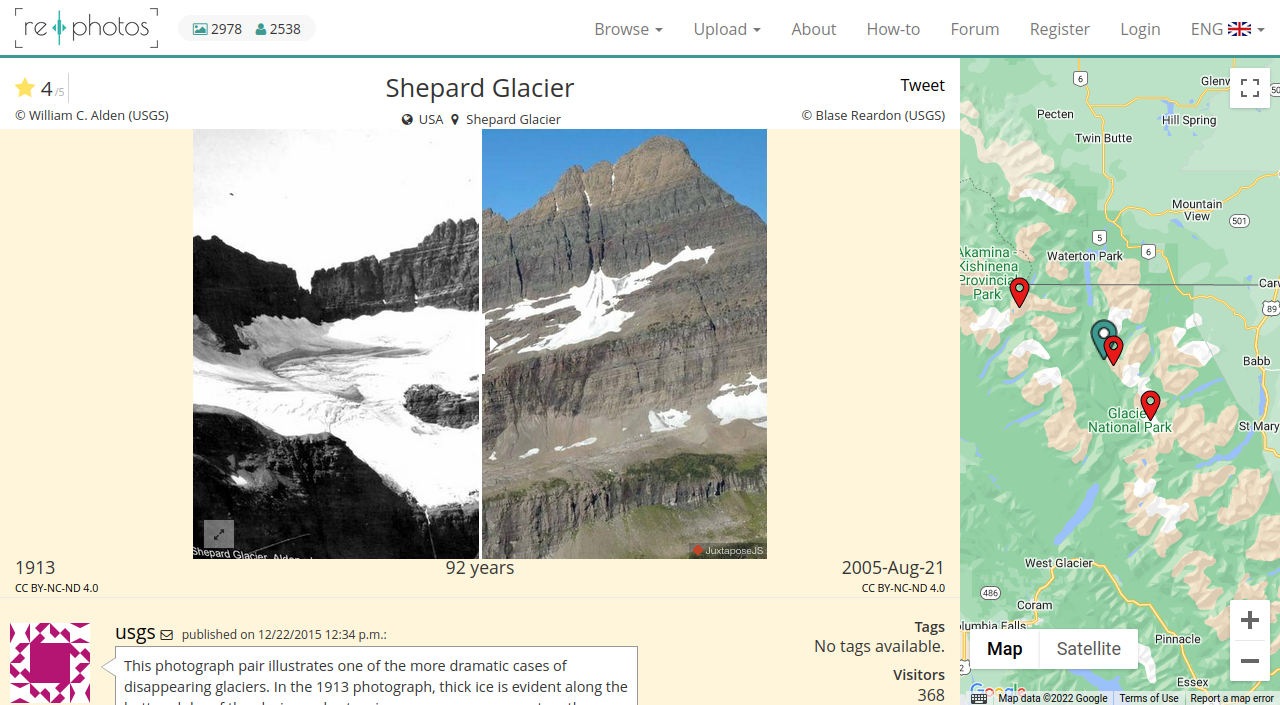}
	\caption{A rephotographic compilation on \rp. Users use the slider in the 
	middle to reveal more 
	of the old or new image. Meta information is displayed around the 
	compilation, including a description, comments, and date. A map section on 
	the right shows the location of the compilation together with neighbouring 
	compilations.}
	\label{fig:port1}
\end{figure}

Currently we have \rp\footnotemark[\value{footnote}]
a web portal covering the complete collaborative
rephotography pipeline. Users  of the portal can upload and georeference template images
which can be searched for with the help of meta information or on a map.
These template images can then be rephotographed and uploaded by the user themselves or 
others. During the upload process meta information is added by the user and the rephotograph
is aligned to the template image with interactive image registration techniques\ci{schaffland2019}.
The completed rephotographic compilations are then added to the database. 
The compilations can be displayed on a map or as a list filtered by meta 
information as shown in Figure\re{fig:port2}.
Meta information includes location, date, description, rating, comments, license, etc.
The compilations are displayed with an interactive slider revealing either more of the
template or the rephotograph as displayed in Figure\re{fig:port1}.
\rp checks many boxes for a mass rephotography portal: It offers a central storage
location of images and metadata as well as methods to filter, access, visualize, and share them.
Together with interactive image registration it is a very good tool for individual
rephotography. Nonetheless, for mass rephotography additional requirements have to be
met. These are presented in the next section. 

Another aspect to keep in mind is the heterogeneous vocabulary in rephotography. 
Rephotographic projects rarely have the term rephotography in their title. 
The norm is to have no terms related to rephotography or terms like ``then and now'' and ``before and after''. For example, one of the first rephotographic projects
\emph{Anatomy of Foliage} created by Edward Fox in 1865/1866 contains no rephotography 
related terms in its title or description.
If nothing else, naming a rephotographic project appropriately, is paramount
to retrieve it reliably and possibly integrate it into a future mass 
rephotography portal.

\section{Where We Need to Go}
\label{where-we-need-to-go}

\subsection{Technique}
\label{technique}
For mass rephotography, we need centralized storage and a web portal
with similar features as \rp and additional features focusing on automatization
and integration. 
We need automatic image registration methods not only 
to align existing template images with existing rephotographs
but also to guide rephotographers to the camera position of the 
template image while rephotographing.
With the recent developments
in machine learning, this seems possible within the next years\ci{maiwald2021}.

Batch upload for georeferenced template images is imperative,
as argued by Robert Campbell in the first contribution.
Without a vast collection of georeferenced template images no 
mass rephotography is possible. Institutions sitting on a
treasure of georeferenced images suitable for rephotography 
need an efficient way to integrate their treasure into the
mass rephotography database. Consequently, methods for mass  or 
possible automatic georeferencing are required, too.

Batch upload for rephotographic collections is another
requirement to efficiently upload existing collections. One solution
would be the upload of an archive of all images together with a database 
file containing the meta information of all images. This database file could
be exported from a \gls{gis}, which was used for georeferencing or managing
the project, and integrated into the mass rephotography database.
Another solution would be an interactive batch upload method integrated in
the portal, where users can upload several images
in one go and add meta information afterwards simultaneously for several 
images.\footnote{Wikimedia Commons, for example, has both locally installable 
programs to prepare image for batch upload and online methods integrated 
into the website.}

Another requirement of mass rephotography is the development of smartphone app, integrated into the portal
In the app, template images near the user's location are displayed,
which can be selected for rephotography. The app then navigates 
the user to the correct camera position and helps align the 
orientation of the smartphone camera with the orientation of the
historic camera similar as it was previously done with a SLR camera and 
a laptop\ci{bae2010computational}. Date and location information can 
be added automatically while the user  provides additional meta information in a standardized format 
before the rephotograph is uploaded and integrated into the mass rephotography database.
The app can also be used to gamify rephotography and to share 
rephotographs to social media to increase public engagement in 
rephotography.

A vital part of the mass rephotography portal are pages dedicated
to projects, groups, and institutions. These pages feature not only 
template images and rephotographs belonging to this project 
but also give room to present 
the group or institution itself as well as background information,
goals, and research results of the project. This would give an incentive
to upload rephotographic collections to the mass rephotography portal
and use it for research and projects employing rephotography.

Persistence and accessibility are also key factors to conserve
our rephotographic heritage. Persistent document identifiers for
the rephotographic compilations and collections are a requirement
as well as a long-term storage solution for the compilations.
Storage, presentation, and search of meta information is another
key aspect. This does not only include ``hard'' meta information
like location and date but also ``soft'' meta information, like the
human historical evidence discussed by Dani Inkpen in his contribution.

Appealing presentation and visualization formats are an additional
requirement. Traditionally, rephotographic compilations were 
printed with old and new images next to each other. For 
digital rephotography there
are many more options available from alpha blending over interactive sliders
and time reveal\footnote{A technique in which one image is overlayed 
over the other. The underlying image is revealed around the cursor position and
follows the mouse movements of the user. First use on a DVD of the ``Third views, second sights'' project\ci{klett2004}.} to videos and animations like GIFs and APNGs.
All have different advantages
and disadvantages depending on the application\ci{schaffland2022}. Thus, the mass 
rephotography portal should offer a selection of these, combined
with an option to share them to social media.

Overall, we need to work through a long list of technical tasks to create
a mass rephotography portal, but with \rp we can start on a solid
foundation.

\subsection{Mindset}
\label{mindset}
Besides new techniques, mass rephotography warrants a new mindset.
\emph{Sharing in caring} holds true for template images,
rephotographs, and complete collections. If we want to preserve our rephotographic heritage,
these images can no longer remain in private archives but have
to be made available to the public under a license allowing 
use in (research) projects, publications and for future rephotography. The same holds for completed
research projects using rephotography. If their data is
public, others can use this data in their research, read about the project
on the project pages discussed above, and use the rephotographs
as templates for future rephotography.

\emph{Location, location, location}---without it the expense of
rephotography rises immensely. While many images can be 
georeferenced by experts (read that as expensive) at a later point in time, 
modern digital
and smartphone cameras can directly embed location information in 
the metadata of the image. Hence, it is essential to enable this 
feature and also share the images together with this location metadata. 
This is true for new template images but also for rephotographs,
helping to verify the location information of the template images,
and simplifying future rephotography.

\subsection{Quantity, Quality, and Relevance}
\label{quantity-quality-and-relevance}
The quantity of mass rephotography inevitably entails questions regarding
quality and relevance. Which content is relevant for mass rephotography?
Templates taken by well-known photographers of well-known places? Template
images taken before the turn of the century and of relevance for landscape
change research? All Google Street View images taken at different points in
times?\footnote{For some locations different recording dates can be selected 
in Google Street view resulting in coarsely registered panoramas.} And regarding quality: What are the quality standards? Registration quality, 
image quality, or image resolution? Who ensures that quality and relevance 
criteria are met and who is eligible to specify these criteria? Clearly 
this can only be a joint effort inspired by platforms like Wikimedia
Commons.

\section{Alternatives}
\label{alternatives}

\subsection{Reinventing the Wheel}
\label{reinventing-the-wheel}
Certainly not an alternative is reinventing the wheel over and over
again by developing yet another rephotographic project page dedicated
to one single project or region---just as nobody would develop
a new clone of YouTube to share their videos.
Besides the high development costs, these project
pages must be maintained, which requires money and time. If one 
or both are not available, the project page vanishes and the 
rephotographs are lost. The ``Alpines of the Americas''
\footnote{\url{https://web.archive.org/web/20180224184356/http://alpineamericas.com/}}
project is one example, 
``The Repeat Photography Project''
\footnote{\url{https://web.archive.org/web/20190221193619/http://www.repeatphotography.org/intro/}} 
of the Forest History Society another. These two and others serve
as example that we need a persistent and accessible storage solution 
for our digital rephotographic heritage. Further, like their analog 
counterparts, the project pages are not easy to find, if they do 
not include terms related to rephotography in their title or description.

\subsection{Centralized Search}
\label{centralized-search}
Combining decentralized project pages with a dedicated centralized search engine for rephotographs (or an option to search for rephotographs embedded into a standard search engine) would be a step in the right direction. This would give owners more control over their rephotographs, while still making them accessible through a central interface. It would require methods to automatically search, index, and identify rephotographs, possibly requiring an extension of an existing image file format, to encompass two or more images with meta information---of course such a file format would be beneficial for rephotography in general. However, this alternative suffers from the same disadvantage as solely reinventing the wheel. If one of the decentralized project pages goes offline, the rephotographs on this server are lost.

\section{Conclusion}
\label{conclusion}
While \rp is a solid foundation for a mass rephotography portal, the 
list of tasks is long. Technical challenges like 
image registration, batch processing of template image and rephotographs, 
persistent storage, and specialized file formats
must be addressed and a change of mindset towards a more liberal approach
to sharing georeferenced templates and rephotographs together with 
meta information is required.
Alternatives like separate project pages are not optimal for mass rephotography,
even if they are combined with a centralized search engine. When funding or
time for maintenance runs out these pages vanish together with their templates 
and rephotographs. A mass rephotography portal with centralized storage combining
and extending functionality of \rp  and Wikimedia Commons seems to be the best option.

\bibliographystyle{ACM-Reference-Format}
\bibliography{rePhotos.bib}


\begin{thebibliography}{5}


\ifx \showCODEN    \undefined \def \showCODEN     #1{\unskip}     \fi
\ifx \showDOI      \undefined \def \showDOI       #1{#1}\fi
\ifx \showISBNx    \undefined \def \showISBNx     #1{\unskip}     \fi
\ifx \showISBNxiii \undefined \def \showISBNxiii  #1{\unskip}     \fi
\ifx \showISSN     \undefined \def \showISSN      #1{\unskip}     \fi
\ifx \showLCCN     \undefined \def \showLCCN      #1{\unskip}     \fi
\ifx \shownote     \undefined \def \shownote      #1{#1}          \fi
\ifx \showarticletitle \undefined \def \showarticletitle #1{#1}   \fi
\ifx \showURL      \undefined \def \showURL       {\relax}        \fi
\providecommand\bibfield[2]{#2}
\providecommand\bibinfo[2]{#2}
\providecommand\natexlab[1]{#1}
\providecommand\showeprint[2][]{arXiv:#2}

\bibitem[Bae et~al\mbox{.}(2010)]%
        {bae2010computational}
\bibfield{author}{\bibinfo{person}{Soonmin Bae}, \bibinfo{person}{Aseem
  Agarwala}, {and} \bibinfo{person}{Fr{\'e}do Durand}.}
  \bibinfo{year}{2010}\natexlab{}.
\newblock \showarticletitle{Computational Rephotography}.
\newblock \bibinfo{journal}{\emph{ACM Trans. Graph}} \bibinfo{volume}{29},
  \bibinfo{number}{3} (\bibinfo{year}{2010}), \bibinfo{pages}{1--15}.
\newblock
\urldef\tempurl%
\url{https://doi.org/10.1145/1805964.1805968}
\showDOI{\tempurl}


\bibitem[Klett(2004)]%
        {klett2004}
\bibfield{author}{\bibinfo{person}{Mark Klett}.}
  \bibinfo{year}{2004}\natexlab{}.
\newblock \bibinfo{booktitle}{\emph{Third views, second sights: a
  rephotographic survey of the American West}}.
\newblock \bibinfo{publisher}{Museum of New Mexico Press},
  \bibinfo{address}{Santa Fe}.
\newblock


\bibitem[Maiwald et~al\mbox{.}(2021)]%
        {maiwald2021}
\bibfield{author}{\bibinfo{person}{Ferdinand Maiwald},
  \bibinfo{person}{Christoph Lehmann}, {and} \bibinfo{person}{Taras Lazariv}.}
  \bibinfo{year}{2021}\natexlab{}.
\newblock \showarticletitle{Fully Automated Pose Estimation of Historical
  Images in the Context of 4D Geographic Information Systems Utilizing Machine
  Learning Methods}.
\newblock \bibinfo{journal}{\emph{ISPRS International Journal of
  Geo-Information}} \bibinfo{volume}{10}, \bibinfo{number}{11}
  (\bibinfo{year}{2021}), \bibinfo{pages}{1--25}.
\newblock
\urldef\tempurl%
\url{https://doi.org/10.3390/ijgi10110748}
\showDOI{\tempurl}


\bibitem[Schaffland and Heidemann(2022)]%
        {schaffland2022}
\bibfield{author}{\bibinfo{person}{Axel Schaffland} {and}
  \bibinfo{person}{Gunther Heidemann}.} \bibinfo{year}{2022}\natexlab{}.
\newblock \showarticletitle{Heritage and Repeat Photography: Techniques,
  Management, Applications, and Publications}.
\newblock \bibinfo{journal}{\emph{Heritage}} \bibinfo{volume}{5},
  \bibinfo{number}{4} (\bibinfo{year}{2022}), \bibinfo{pages}{4267--4305}.
\newblock
\urldef\tempurl%
\url{https://doi.org/10.3390/heritage5040220}
\showDOI{\tempurl}


\bibitem[Schaffland et~al\mbox{.}(2019)]%
        {schaffland2019}
\bibfield{author}{\bibinfo{person}{Axel Schaffland}, \bibinfo{person}{Oliver
  Vornberger}, {and} \bibinfo{person}{Gunther Heidemann}.}
  \bibinfo{year}{2019}\natexlab{}.
\newblock \showarticletitle{An Interactive Web Application for the Creation,
  Organization, and Visualization of Repeat Photographs}. In
  \bibinfo{booktitle}{\emph{1st Workshop on Structuring and Understanding of
  Multimedia heritAge Contents (SUMAC {`}19)}}. \bibinfo{publisher}{ACM},
  \bibinfo{address}{New York, NY, USA}, \bibinfo{pages}{47--54}.
\newblock
\urldef\tempurl%
\url{https://doi.org/10.1145/3347317.3357247}
\showDOI{\tempurl}


\end{thebibliography}

\end{document}